\documentclass{article}

\usepackage{microtype}
\usepackage{graphicx}
\usepackage{subfigure}
\usepackage{booktabs} 

\usepackage{hyperref}

\usepackage[accepted]{icml2024}

\usepackage{amsmath}
\usepackage{amssymb}
\usepackage{mathtools}
\usepackage{amsthm}

\usepackage{multirow}
\usepackage{makecell}
\usepackage{makecell}

\usepackage[capitalize,noabbrev]{cleveref}

\theoremstyle{plain}

\theoremstyle{definition}

\theoremstyle{remark}

\usepackage[textsize=tiny]{todonotes}

\icmltitlerunning{Evaluating and Analyzing Relationship Hallucinations in LVLMs}

\begin{document}

\twocolumn[
\icmltitle{Evaluating and Analyzing Relationship Hallucinations \\in Large Vision-Language Models}



\icmlsetsymbol{equal}{*}

\begin{icmlauthorlist}
\icmlauthor{Mingrui Wu}{mac}
\icmlauthor{Jiayi Ji}{mac}
\icmlauthor{Oucheng Huang}{mac}  
\icmlauthor{Jiale Li}{mac}
\icmlauthor{Yuhang Wu}{mac}
\icmlauthor{Xiaoshuai Sun}{mac}
\icmlauthor{Rongrong Ji}{mac}
\end{icmlauthorlist}

\icmlaffiliation{mac}{Key Laboratory of Multimedia Trusted Perception and Efficient Computing, Ministry of Education of China, Xiamen University, 361005, P.R. China}

\icmlcorrespondingauthor{Jiayi Ji}{jjyxmu@gmail.com}

\icmlkeywords{LVLMs, Relationship Hallucination}

\vskip 0.3in
]



\printAffiliationsAndNotice{}  

\begin{abstract}
The issue of hallucinations is a prevalent concern in existing Large Vision-Language Models (LVLMs). Previous efforts have primarily focused on investigating object hallucinations, which can be easily alleviated by introducing object detectors. However, these efforts neglect hallucinations in inter-object relationships, which is essential for visual comprehension. In this work, we introduce R-Bench, a novel benchmark for evaluating Vision Relationship Hallucination. R-Bench features image-level questions that focus on the existence of relationships and instance-level questions that assess local visual comprehension. We identify three types of relationship co-occurrences that lead to hallucinations: relationship-relationship, subject-relationship, and relationship-object. The visual instruction tuning dataset's long-tail distribution significantly impacts LVLMs' understanding of visual relationships. Furthermore, our analysis reveals that current LVLMs tend to disregard visual content and overly rely on the common sense knowledge of Large Language Models. They also struggle with reasoning about spatial relationships based on contextual information. Github: \url{https://github.com/mrwu-mac/R-Bench}.

\end{abstract}

\section{Introduction}
\label{intro}
Recently, large language models (LLMs) such as GPT-4~\cite{openai2023gpt4} and Llama~\cite{touvron2023Llama} have demonstrated significant capabilities in addressing a broad spectrum of human-generated questions. The success of these models has spurred researchers to explore the use of LLMs in conjunction with visual inputs, leading to the development of various large vision-language models (LVLMs)~\cite{li2023blip, liu2023visual, dai2305instructblip, ye2023mplug}. These endeavors typically involve methods like visual language pretraining~\cite{li2023blip} or visual instruction tuning~\cite{liu2023visual}, aimed at integrating pre-trained visual encoders with LLMs to enhance their understanding of visual contexts.

\begin{figure*}[ht]
\vskip 0.1in
\begin{center}
\centerline{\includegraphics[width=0.97\textwidth]{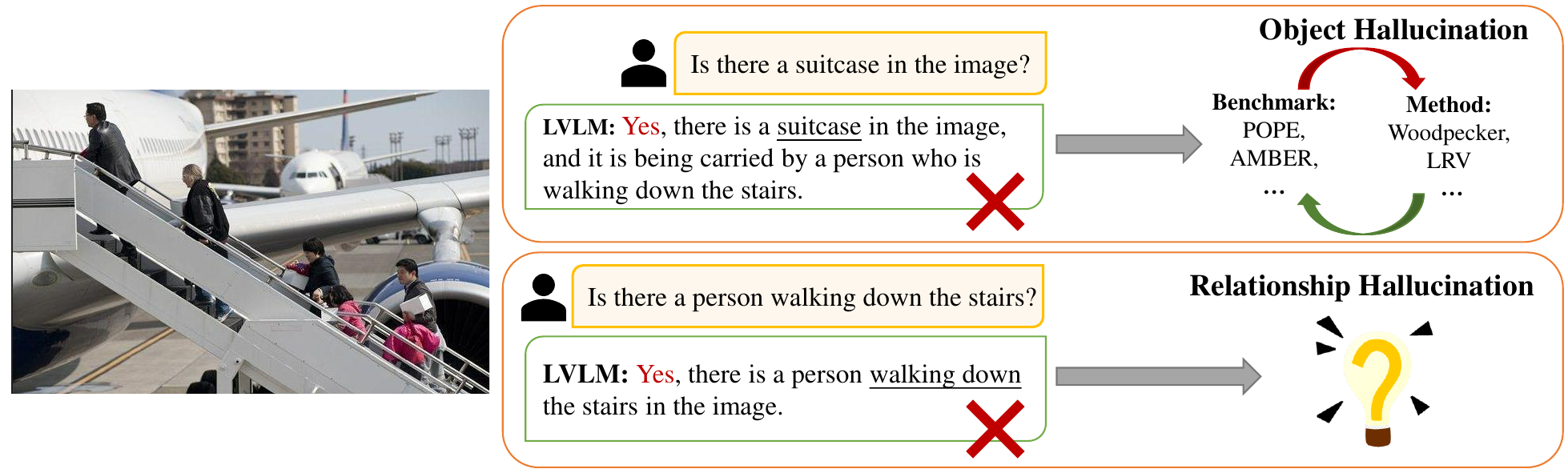}}
\caption{The object hallucination and relationship hallucination in Large Vision-Language Models. While substantial research has addressed object hallucinations in LVLMs, the issue of relationship hallucinations remains under-explored.}
\label{fig:intro}
\end{center}
\vskip -0.3in
\end{figure*}

However, despite their impressive performance, a significant challenge for these models is the unavoidable issue of hallucinations. Existing LVLMs often tend to generate responses that are inconsistent with the content of the images. This issue is particularly critical for LVLMs, which are expected to accurately comprehend images and produce answers consistent with the content of the visual input. While prior research has delved into evaluating object hallucinations~\cite{li2023evaluating}, offering mitigation strategies through the object detection~\cite{yin2023woodpecker} or segmentation models~\cite{wu2022difnet, chen2023mitigating}, there exists a notable gap in addressing hallucinations related to inter-object relationships, as shown in Figure~\ref{fig:intro}. The latter, compared to object hallucinations, better reflects the LVLM's capacity to comprehend the intricacies of visual scenes. 
Currently, there is a shortage of comprehensive and rigorous benchmarks, to address these relationship hallucinations.

In this study, we introduce a novel Relationship Hallucination Benchmark (R-Bench) designed specifically for assessing relationship hallucinations in LVLMs. This benchmark comprises image-level and instance-level questions, labeled as 'Yes' or 'No', similar to the POPE evaluation~\cite{li2023evaluating}. Image-level questions assess the existence of relationships in the image, while instance-level questions focus on specific object relationships, indicated by color bounding boxes or masks. The instance-level questions showcase the local visual understanding ability of LVLMs, adaptable to existing models without requiring retraining.

For the benchmark, we employ a combination of automatic generation by the Large Language Model (LLM) and manual curation. To ensure the benchmark's integrity, it is based on the nocaps validation set~\cite{agrawal2019nocaps}, preventing overlap with the pre-trained data of LVLMs. The construction process involves parsing all COCO captions to create a comprehensive relationship set. For each image in the nocaps dataset, we parse the provided captions into relationship triplets, which are then aligned with the relationship set to establish a set of relationship seeds. Leveraging GroundingDINO, we identify significant objects with bounding boxes. Subsequently, we create prompts based on nocaps captions, relationship seeds, and bounding boxes, which are then fed into LLM to generate both image-level and instance-level questions. Finally, after rigorous manual selection, we have established R-Bench. The total number of questions retained after the filtering process amounts to 11,651, of which 7,883 are image-level questions, and 3,768 are instance-level questions.
%

We assess various recently popular LVLMs on our R-Bench and present our findings as follows:
1) Relationship hallucinations in LVLMs are more severe than object hallucinations, mainly due to the long-tail distribution between relationships and objects in the training data.
2) Relationship hallucinations often emerge from the co-occurrence patterns among relationships, specifically relationship-relationship, subject-relationship, and relationship-object.
3) Employing fine-grained image-text alignment could potentially mitigate hallucinations.
Additionally, our analysis of counterfactual and illusion relationship hallucinations, based on web-collected images, reveals:
4) Existing LVLMs often overlook visual content, relying on LLM's common sense for predictions.
5) LVLMs struggle to reason about spatial relationships based on context.
We aspire that our findings will stimulate the community to explore innovative solutions for mitigating relationship hallucinations in LVLMs.

\begin{figure*}[ht]
\vskip 0.2in
\begin{center}
\centerline{\includegraphics[width=0.99\textwidth]{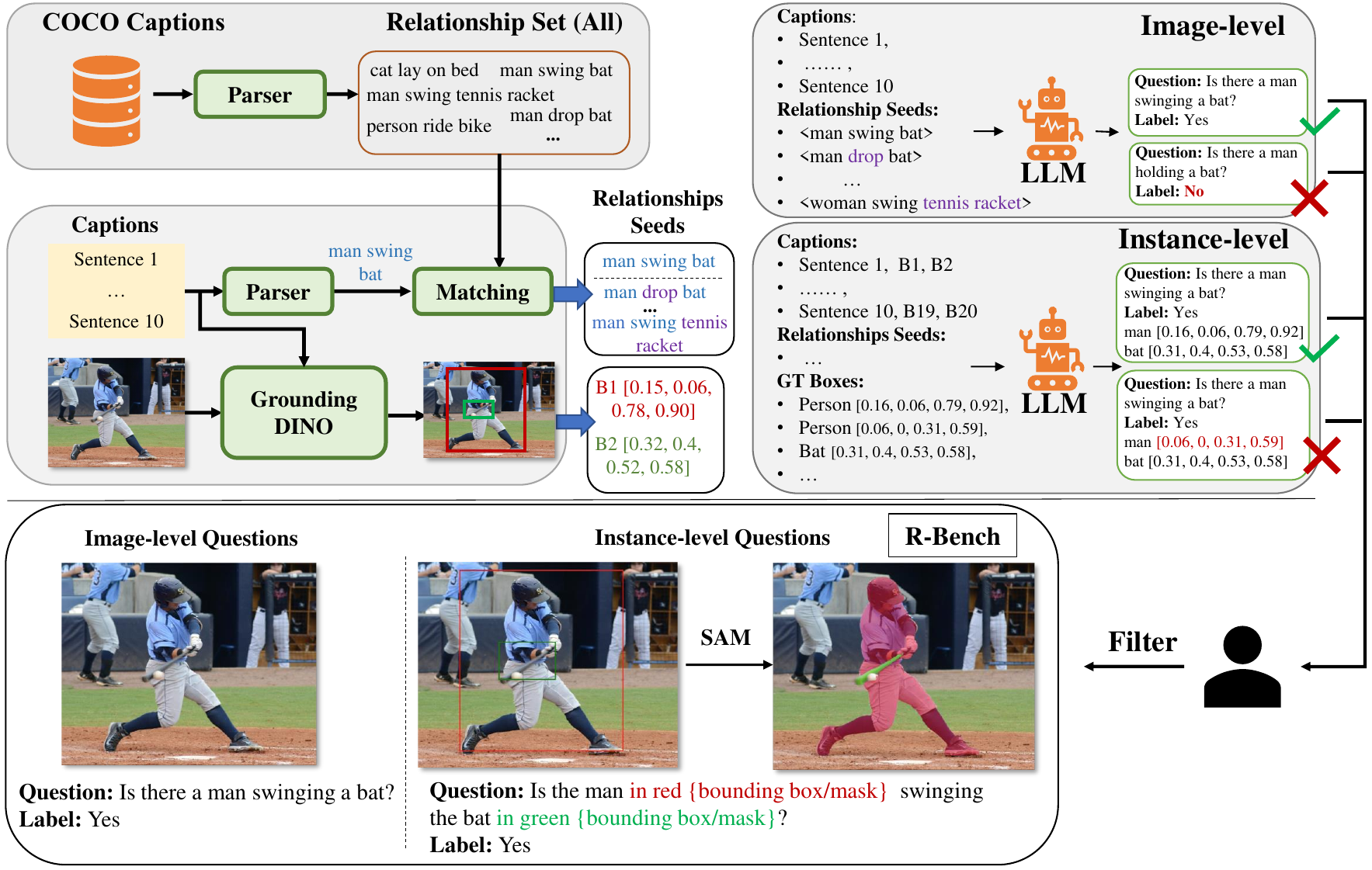}}
\caption{Our pipeline generates image-level and instance-level questions. First, we parse all COCO captions into a relationship set. Given a nocaps image, we parse its corresponding captions into relationship triplets and match these with the relationship set to obtain a set of relationship seeds. Using GroundingDINO, we identify significant objects with bounding boxes. We then create two types of prompts based on the nocaps captions, relationship seeds, and bounding boxes. Finally, we feed these prompts into an LLM to generate image-level and instance-level questions. Additionally, we carefully filter out noisy questions to create the refined R-Bench.}
\label{fig:archi}
\end{center}
\vskip -0.2in
\end{figure*}

\section{Related Work}
\subsection{Large Vision-Language Model}
Motivated by the achievements of Large Language Models (LLMs)~\cite{openai2023gpt4, touvron2023Llama}, there is a current trend among researchers to develop a diverse array of Large Vision-Language Models (LVLMs)~\cite{li2023blip, liu2023visual, dai2305instructblip, ye2023mplug, li2023otter, chen2023shikra, 2023internlm, du2022glm, gao2023Llamaadapterv2, Qwen-VL, luo2023cheap}. These LVLMs typically consist of a visual encoder, a language decoder, and an image-text alignment module, in which the visual encoder and the language decoder are usually from pre-trained models, such as CLIP~\cite{radford2021learning}, Llama~\cite{touvron2023Llama} and Vicuna~\cite{vicuna2023}. The image-text alignment module usually is trained on image-text pairs and finetuned through visual instruction tuning to achieve powerful visual conversation ability. 
However, the training data of LVLMs usually lean towards an intermediate understanding with a focus on individual objects, neglecting the crucial visual comprehension skill of inter-object relationships.

\subsection{Hallucination in LVLMs}
Presently, all LVLMs experience visual hallucination issues, often producing answers inconsistent with image content. Unlike LLMs, these visual hallucinations in LVLMs pose potential real-world harm, impacting their practical applications. 
There are lots of benchmarks~\cite{lu2024mathvista,yue2023mmmu,kembhavi2016diagram,fu2023mme,liu2023mmbench,li2023seed2,yu2023mm,wu2023q,guan2023hallusionbench,masry-etal-2022-chartqa, jing2023faithscore, cheng2023knvqa, villa2023behind, wang2023evaluation, wang2023llm} for evaluating LVLMs. However, the hallucination benchmark is still under-explored. 
Previous research~\cite{li2023evaluating} has focused on exploring object hallucinations, which indicate the presence of objects in the expressions generated by LVLMs that either do not exist in the image or are inconsistent with the image content. Such hallucination can be alleviated by incorporating visual models, such as object detection models~\cite{yin2023woodpecker}, segmentation models~\cite{wu2022difnet, chen2023mitigating} or other methods~\cite{liu2023mitigating,liu2023mitigating, li2023silkie, chen2023dress, gunjal2023detecting, wang2024mitigating, hu2023ciem, zhai2023halle, zhou2023analyzing, tang2023generalization, zhao2023beyond, leng2023mitigating, lu2023lyrics, zhang2021rstnet,wu2023end,wu2024toward}. Some studies~\cite{chen2023mitigating, villa2023behind} have begun to explore object attributes or relative positional relationships between objects hallucinations in LVLMs. Yet, there is a shortage of comprehensive and rigorous benchmarks and lacks effective analysis of hallucinations in inter-object relationships. This paper introduces a novel relationship hallucination benchmark using the nocaps validation set, conducting a thorough assessment to enhance LVLMs' image comprehension capabilities.

\section{Relationship Hallucination Benchmark}
In this section, we describe our Relationship Hallucination Benchmark (R-Bench). Although the pioneering efforts of benchmarks~\cite{chen2023mitigating, wang2023llm} delve into the relationship hallucination, yet inadvertently overlook potential issues related to data leakage. 
These benchmarks are typically constructed directly from the overlapped subset of COCO~\cite{lin2014microsoft} and Visual Gnome~\cite{krishna2017visual}. However, it is noteworthy that the Visual Gnome dataset has been extensively employed in pre-training or visual instruction tuning across the majority of LVLMs~\cite{liu2023improved, li2023blip, ye2023mplug2}. In light of this, we introduce a novel relationship hallucination benchmark, carefully constructed using data from the validation set of nocaps~\cite{agrawal2019nocaps}. Notably, nocaps primarily serves as a captioning evaluator for LVLMs~\cite{dai2305instructblip, li2023blip}. 

\subsection{Benchmark Construction}
We construct two types of benchmarks, the image-level benchmark which focuses on the existence of relationships, and the instance-level benchmark
which assesses local visual comprehension.  The pipeline is shown in Figure~\ref{fig:archi}.

\textbf{Image-level Benchmark.}
We construct questions with a yes or no response, rendering our benchmark potentially compatible with the existing object hallucination benchmark~\cite{li2023evaluating}. 
Additionally, this simple binary classification problem demonstrates better stability compared to instruction-based methods~\cite{li2023evaluating}.
We adopt an LLM to generate the questions based on a pre-extracted set of relationships from COCO captions.
This approach allows us to analyze the hallucinations of LVLMs more effectively, as most LVLMs exhibit the same bias as the COCO caption data distribution due to their visual instruction tuning data being derived from COCO captions.

Specifically, we first apply a scene graph parser~\cite{li-etal-2023-factual} to extract relationship triplets from COCO captions. Subsequently, leveraging the LLM, we generate `Yes' or `No' questions based on both nocaps captions and the pre-extracted set of relationships.
To manage the prompt length effectively, we present only image-related relationships in the prompt for each image. This involves parsing nocaps captions into relationship triplets using the scene graph parser and forming relationship seeds by matching relationships in the set that overlap with any two elements of the current triplet.
Additionally, we guide the LLM to concentrate on generating questions that emphasize relationships and filtering out nonsensical negative questions.

\begin{table}[t]
\caption{The reference object accuracies. We report box accuracy and mask accuracy respectively. The result is for reference only due to ignoring the different words with the same semantics, this can lead to lower accuracy.}
\label{tab:box}
\vskip 0.15in
\begin{center}
\begin{small}
\begin{sc}
\begin{tabular}{lcc}
\toprule
Model & box acc. & mask acc. \\
\midrule
LLaVA-1.5    & 71.04 & 82.60 \\
InstrutBLIP & 72.45 &	77.29 \\
mPLUG-Owl2 & 70.71 & 75.74 \\
Qwen-VL & 76.25 & 70.66 \\
\bottomrule
\end{tabular}
\end{sc}
\end{small}
\end{center}
\vskip -0.1in
\end{table}

\textbf{Instance-level Benchmark.}
The trend in recent LVLMs~\cite{Qwen-VL,zhang2023gpt4roi} is increasing towards incorporating additional inputs,  such as bounding boxes or masks, to enhance local comprehension capabilities.
Additionally, prioritizing the identification of relationships between specified objects aligns more closely with practical demands. To address this, we introduce an instance-level benchmark to assess the relationship between reference objects, demanding LVLMs to possess local visual comprehension. This is achieved by attaching colored bounding boxes or masks to the objects mentioned in the question and visually representing these colored elements on the image. 

\begin{table}[t]
\caption{The data statics of our benchmark after artificial filtering. The `Pos' and `Neg' denote the number of positive and negative questions. The `Obj' and `Rel' denote categories number of objects and relationship (Relationship between subject and object, rather than the relationship triplet) respectively. The total number of the `Image', `Obj', and `Rel' minus the overlap of the ones between image-level and instance-level.}
\label{tab:sta}
\vskip 0.15in
\begin{center}
\begin{small}
\begin{sc}
\begin{tabular}{lccccc}
\toprule
\multirow{2}*{Type} & \multicolumn{2}{c}{Question} & \multirow{2}*{Image} & \multirow{2}*{Obj} & \multirow{2}*{Rel} \\
\cline{2-3}
 ~ & Pos & Neg & ~ & ~ & ~ \\
\midrule
Image    & 5,134 & 2,749 & 3,657  & 2,602 & 587  \\
Instance & 2,896 & 872 & 2,645 & 2,317 & 511 \\
\midrule
Total & 8,030 & 3,621 & 4,034 & 3,514 & 791 \\
\bottomrule
\end{tabular}
\end{sc}
\end{small}
\end{center}
\vskip -0.15in
\end{table}

\begin{table*}[t]
\caption{Results of LVLMs on our R-Bench. We compute average scores of 5 random subsets, and each subset has 1:1 pos-neg questions. The `box' and `mask' denote types of instance-level questions with bounding box and mask respectively.}
\label{tab:result}
\vskip 0.15in
\begin{center}
\begin{small}
\begin{sc}
\begin{tabular}{llccc|c|c}
\toprule
Type & Model & Accuracy & Precision & Recall & F1 Score & Yes \\
\midrule
\multirow{4}*{Image-level(All)} & LLaVA-1.5   & 71.23 & 64.27 & 96.89 & 77.28 & 76.12 \\
~ & InstrutBLIP & 69.31 & 62.76 & 96.45 & 76.04 & 77.60  \\
~ & mPLUG-Owl2 & 73.66 & 67.60 &  91.84 & 77.88 & 68.60 \\
~ & Qwen-VL & 79.19 & 76.43 & 84.99 & 80.48 & 56.15 \\
\midrule
\multirow{4}*{Image-level(Subset)} & LLaVA-1.5 & 71.30 & 64.31 & 97.11 & 77.38 & 76.32 \\
& InstructBLIP & 69.21 & 62.60 & 97.11 & 76.12 & 78.41 \\
& mPLUG-OWL2 & 74.66 & 68.40 & 92.67 & 78.71 & 68.48 \\
& Qwen-VL & 79.51 & 76.38 & 85.48 & 80.67 & 55.98 \\
\midrule
\multirow{4}*{Box} & LLaVA-1.5    & 53.15 & 51.71 & 95.53 & 67.10 &  92.37  \\
~ & InstrutBLIP & 51.95 & 51.14 & 87.39 & 64.52 & 85.44 \\
~ & mPLUG-Owl2 & 53.90 &  52.38 & 85.67 & 65.01 & 81.77 \\
~ & Qwen-VL & 58.82 & 55.56 & 88.17 & 68.16 & 79.35 \\
\midrule
\multirow{4}*{Mask} & LLaVA-1.5    & 53.44 & 51.89 & 94.50 &  66.99 & 91.06  \\
~ & InstrutBLIP & 55.63 & 53.61 & 83.51 & 65.30 &  77.88   \\
~ & mPLUG-Owl2 & 55.80 & 55.46 & 80.44 & 65.65 & 76.17  \\
~ & Qwen-VL & 59.84 & 57.65 & 74.15 & 64.87 & 64.31  \\
\bottomrule
\end{tabular}
\end{sc}
\end{small}
\end{center}
\vskip -0.1in
\end{table*}

We first verify that existing LVLMs can distinguish reference objects. To assess this ability without considering whether the images have been encountered by LVLMs, we utilize the COCO validation set along with corresponding instance segmentation annotations, incorporating bounding boxes or masks. Since the existing LVLMs cannot detect and identify small objects, we first use COCO captions and GroundingDINO~\cite{liu2023grounding} to extract the significant object area and match it with the ground truth to obtain the accurate object position. Subsequently, we visualize the identified objects by drawing bounding boxes or masks in random red or green colors on the images. To prompt the LVLMs, we use questions like ``What is the object in the \{red/green\} \{bounding box/mask\}?''. The typical response is ``The object in the \{red/green\} bounding box is a \{object\}'', allowing us to compute accuracy by matching the label with this response.

It is important to note that this approach may yield lower accuracy due to the complexities introduced by open-set questions, such as potential variations in object names with correct meanings, such as ``bike'' and ``bicycle'' (illustrated in Figure~\ref{fig:ref_obj}). We evaluate several recent LVLMs, and the results are presented in Table~\ref{tab:box} for reference. Notably, all LVLMs demonstrate relatively high accuracy, indicating their proficiency in discriminating reference objects. Moreover, most LVLMs exhibit higher accuracy in mask-based object discrimination compared to box-based methods, likely due to the more precise references provided by masks. Notably, Qwen-VL~\cite{Qwen-VL} performs box-level image-text alignment training, thus getting a better performance on box accuracy. 

Next, we generate the questions with bounding boxes and masks. We first feed the nocaps captions into the GroundingDINO to get a set of object bounding boxes, which usually have a stronger correlation with the captions. Then we let LLM generate the yes or no questions based on nocaps captions paired with extracted object bounding boxes, related relationship set, and some ground-truth bounding boxes of the image sourced from OpenImage (as the nocaps images are derived from OpenImage~\cite{kuznetsova2020open}).
The LLM outputs questions containing bounding boxes for the subjects and objects of relationships. We further feed these bounding boxes into the SAM~\cite{kirillov2023segment} to get the corresponding masks. Upon obtaining questions with bounding boxes and masks, we visualize them on the image by incorporating colored bounding boxes or masks. When evaluating LVLMs on this instance-level benchmark, we present images adorned with these visual elements, and the prompts are formatted as ``Is there \{subject\} in the red \{bounding box/mask\} relationship \{object\} in the green \{bounding box/mask\} in the image?".

This method not only enhances the model's ability for fine-grained analysis but also facilitates a direct comparison and integration with prior research.

\subsection{Data Statics}
We obtain a total of 24,897 questions. Due to insufficient image information provided to the LLM, it tends to generate numerous noisy questions. So we employ filters to filter the generated questions. These filters screen out questions if they meet any of the following criteria: (1) the question contains a typographical error; (2) the question is not related to relationships; (3) the answer label is incorrect or cannot be determined; (4) there is a discrepancy between the object described in the question and the object highlighted by the box or mask, among others. Filtering out the noise and ensuring logical and label-correct questions takes approximately two weeks. 

The total number of questions after filtering is 11,651, in which the number of image-level questions is 7,883 with 5,134 positive questions and 2,749 negative questions, the number of instance-level questions is 3,768 with 2,896 positive questions and 872 negative questions, as shown in Table~\ref{tab:sta}. The instance-level with box and mask share the same questions. The total image number is 4,034. The category number of objects in questions is 3,514, of which 3,028 objects appear in COCO captions. The category number of relationships in questions is 791, of which 739 relationships appear in COCO captions. And the relationship types include actions, spatial relationships, relationship existence, and so on. These objects and relationships together form a total of 11,335 relationship triples in the questions, of which 4,941 appear in COCO captions. On average, each question has a unique relationship triplet.
Due to the severe imbalance between positive and negative questions, we randomly select an equal number of both to create a subset for LVLMs evaluation.

\begin{table*}[t]
\caption{Comparing image-level relationship hallucination with image-level object hallucination. We apply POPE to get an object hallucination set on the validation set of the nocaps. 
We construct two corresponding object questions for each relationship question for POPE adversarial, popular and random setting, and report the mean results.}
\label{tab:r_o}
\vskip 0.15in
\begin{center}
\begin{small}
\begin{sc}
\begin{tabular}{llccc|c|c}
\toprule
Type & Model & Accuracy & Precision & Recall & F1 Score & Yes \\
\midrule
\multirow{4}*{Image-Obj} &	LLaVA-1.5 &	78.34 & 71.01 & 97.81 & 82.12 & 69.54 \\
& InstructBLIP & 78.77 & 72.12 & 96.72 & 82.36 & 67.94 \\
& mPLUG-OWL2 & 77.11 & 71.24 & 92.36 & 80.31 & 65.25 \\
& Qwen-VL & 86.57 & 93.29 & 78.99 & 85.50 & 42.42 \\
\midrule
\multirow{4}*{Image-Rel} & LLaVA-1.5 & 71.23 & 64.27 & 96.89 & 77.28 & 76.12 \\
& InstructBLIP & 69.31 & 62.76 & 96.45 & 76.04 & 77.60 \\
& mPLUG-OWL2 & 73.66 & 67.60 & 91.84 & 77.88 & 68.60 \\
& Qwen-VL & 79.19 & 76.43 & 84.99 & 80.48 & 56.15 \\

\bottomrule
\end{tabular}
\end{sc}
\end{small}
\end{center}
\vskip -0.1in
\end{table*}

\begin{figure*}[ht]
\vskip 0.2in
\begin{center}
\centerline{\includegraphics[width=0.99\textwidth]{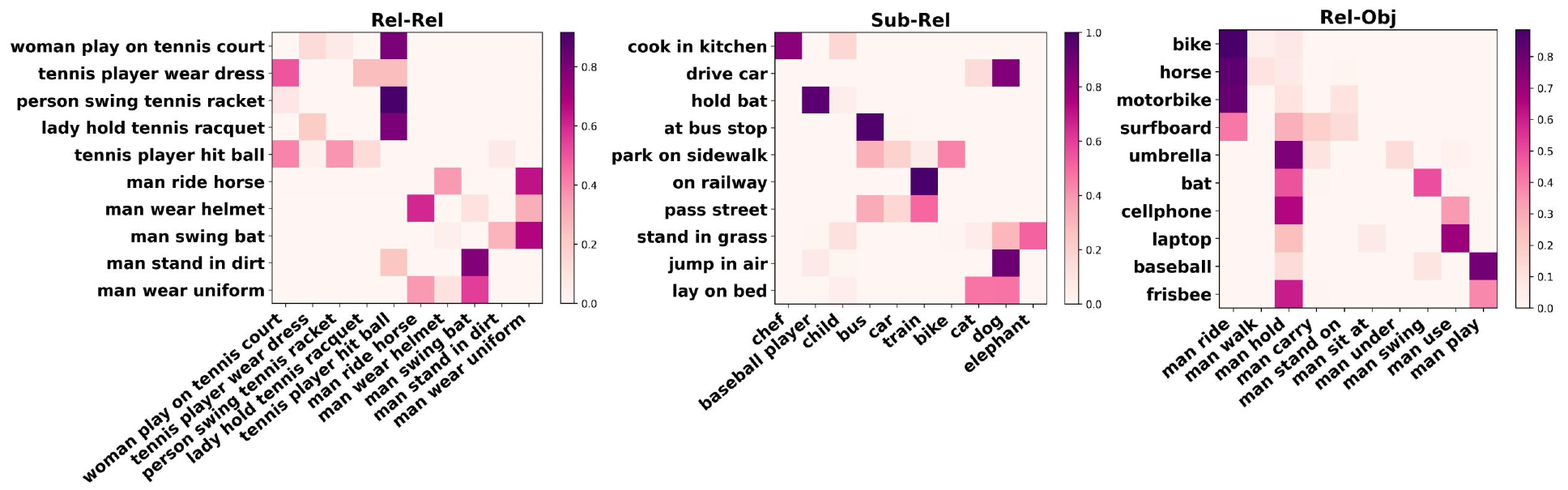}}
\caption{The co-occurrence matrices constructed between relationship-relationship~(left), subject-relationship~(middle), and relationship-object~(right) respectively. The matrices show the conditional probability that an element of the y-axis occurs when another element of the x-axis is happening.}
\label{fig:co}
\end{center}
\vskip -0.2in
\end{figure*}

\begin{figure*}[ht]
\vskip 0.2in
\begin{center}
\centerline{\includegraphics[width=0.95\textwidth]{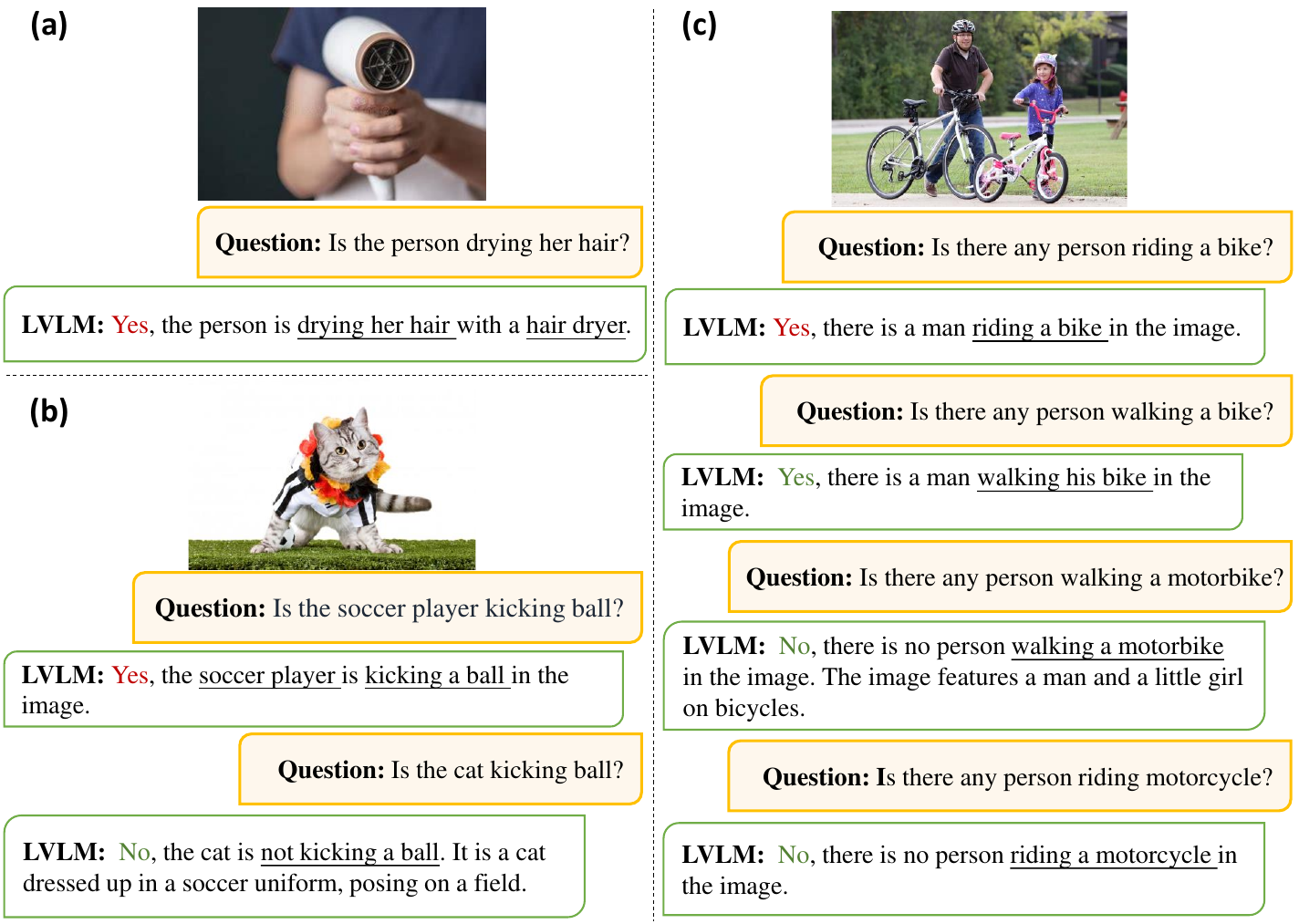}}
\caption{The examples of relationship hallucination which arise for different reasons. The wrong answers are marked in red, the relationships in the answers are underlined, and correct answers are marked in green.}
\label{fig:co3}
\end{center}
\vskip -0.2in
\end{figure*}

\begin{figure}[ht]
\begin{center}
\centerline{\includegraphics[width=0.49\textwidth]{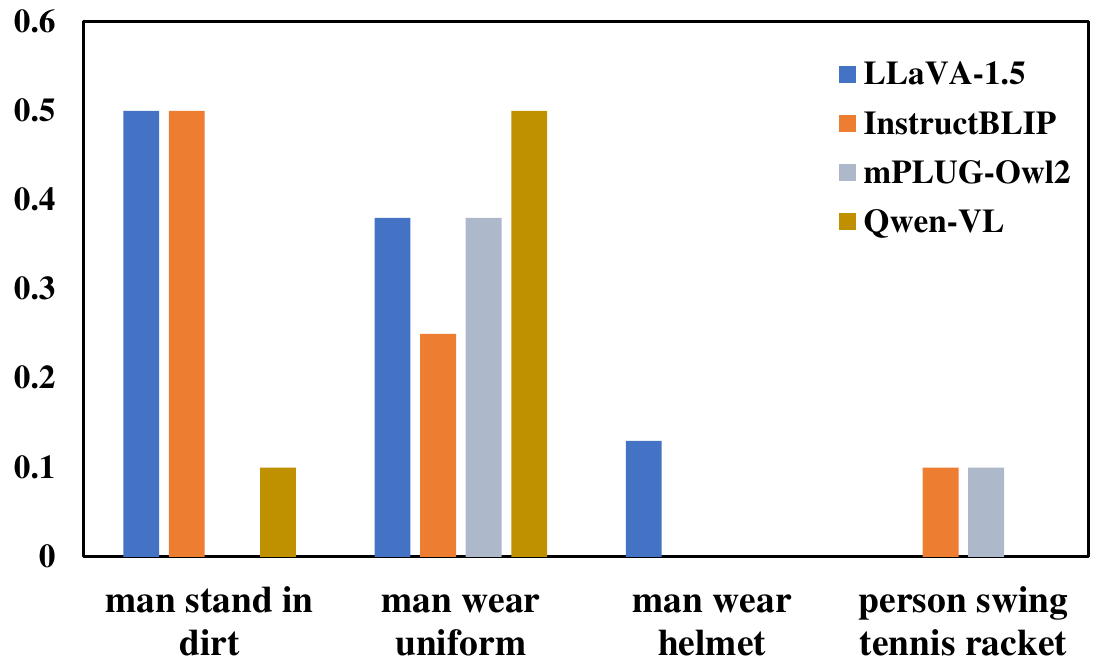}}
\caption{The probability of relationship hallucination when ``man swings bat'' occurs. The co-occurrence frequencies of these relationships with ``man swings bat'' decrease from left to right.}
\label{fig:web1}
\end{center}
\vskip -0.3in
\end{figure}

\begin{figure}[ht]
\begin{center}
\centerline{\includegraphics[width=0.49\textwidth]{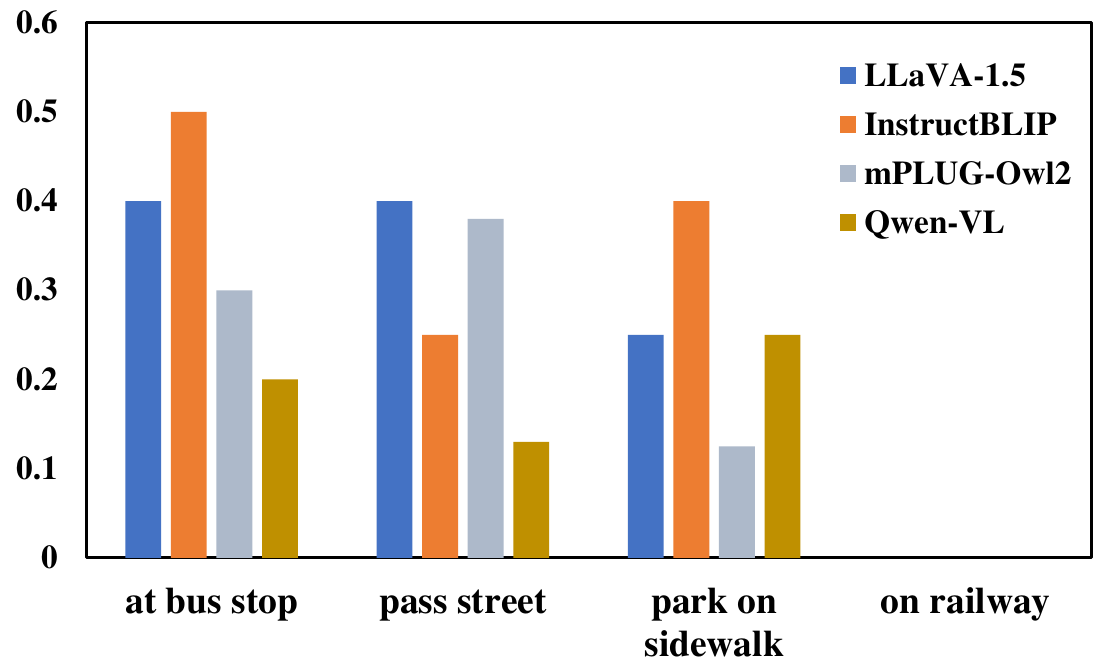}}
\caption{The probability of relationship hallucination when the subject is ``bus''. The co-occurrence frequencies of these relationships with ``bus'' decrease from left to right.}
\label{fig:web2}
\end{center}
\vskip -0.3in
\end{figure}

\begin{figure}[ht]
\begin{center}
\centerline{\includegraphics[width=0.49\textwidth]{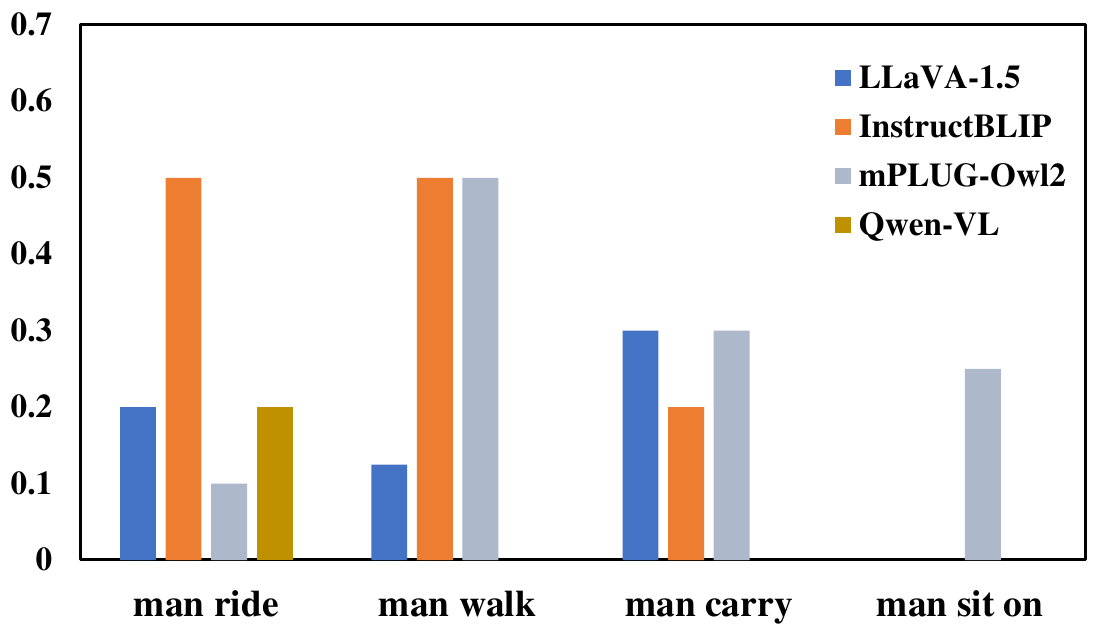}}
\caption{The probability of relationship hallucination when the object is ``bike''. The co-occurrence frequencies of these relationships with ``bike'' decrease from left to right.}
\label{fig:web3}
\end{center}
\vskip -0.3in
\end{figure}

\section{Experiments}
\subsection{Experimental Setup}
\textbf{Datasets and Evaluation Metrics.}
Our benchmark is built upon the validation set of nocaps~\cite{agrawal2019nocaps}, which consists of 4500 images from OpenImage~\cite{kuznetsova2020open} and each image with 10 captions. These images can be divided into 648 in-domain images~(only with COCO classes), 2938 near-domain images~(with COCO and novel classes) and 914 out-domain images~(only with novel classes). We follow the POPE~\cite{li2023evaluating} to use Accuracy, Precision, Recall, F1 score, and Yes ratio as the evaluation metrics.  

\textbf{Implementations.}
We evaluate several recently popular LVLMs: LLaVA-1.5~\cite{liu2023visual,liu2023improved}, InstructBLIP~\cite{dai2305instructblip}, mPLUG-Owl2~\cite{ye2023mplug2}, Qwen-VL~\cite{Qwen-VL}. All models are evaluated on 1 NVIDIA RTX-3090 with 24G memory. And the LLM used to generate questions is Llama2-chat-13B~\cite{touvron2023Llama}. 
More information about models is shown in the Appendix.

\subsection{Evaluating on the R-Bench}
\textbf{Evaluating LVLMs.}
We evaluate several popular LVLMs on our Relationship Hallucination Benchmark (R-Bench) under three settings: image-level, instance-level with bounding box, and instance-level with mask. For each setting, we randomly sample 5 subsets with a 1:1 positive-negative question ratio and compute the average scores of these 5 subsets. Additionally, we evaluate an extra image-level subset using the same set of images as the instance-level subsets, allowing us to more effectively assess the differences in hallucination issues between image-level and instance-level questions for LVLMs.

The results are shown in Table~\ref{tab:result}. In contrast to image-level hallucinations, the LVLMs consistently hallucinate on instance-level settings. Although these LVLMs have a good recognition ability for reference objects as shown in Table~\ref{tab:box}, they all fail to discriminate the relationship between reference objects. We show examples of image-level and instance-level relationship hallucinations in Figure~\ref{fig:ins}, the LVLM can answer image-level questions well, but falls in instance-level questions. We believe that a fine-grained image-text alignment might help improve this.

\textbf{Compare Relationship Hallucination with Object Hallucination.}
We assess the significance of addressing relationship hallucinations in LVLMs by comparing them with object hallucinations. To compare relationship hallucination with object hallucination, we use POPE to get an object hallucination set on the validation set of the nocaps. Specifically, we first perform POPE on the validation set of the nocaps based on the combined object labels of SEEM-based and Ground-Truth to obtain an object hallucination set. 
We construct two corresponding object questions for each relationship question for POPE adversarial, popular and random setting, and report the mean results.

The results are shown in Table~\ref{tab:r_o}. Compared with object hallucinations, existing LVLMs have more serious relationship hallucinations. Although Qwen-VL is an exception, it tends to answer no to ensure precision, which leads to a relatively low recall. Overall, existing LVLMs have more room for improvement in relationship hallucinations than object hallucinations.

\subsection{Analysis of the Causes of Hallucinations}
Analyzing the causes of relationship hallucinations will be more helpful for future work. Object co-occurrence is a major contributor to object hallucinations~\cite{li2023evaluating}. Therefore, we analyze whether relationship co-occurrence also leads to relationship hallucination. We hypothesize three types of relationship co-occurrence that may lead to relationship hallucinations, including co-occurrence between the relationship-relationship, subject-relationship, and relationship-object. Unless otherwise specified, we perform the analysis on LLaVA-1.5.

\textbf{Relationship-Relationship Co-occurrence.}
As the visual instruction tuning data is generated based on different captions, some related relationships will co-occur frequently, as shown in Figure~\ref{fig:co}~(left). Such as, when a ``man swing bat'', the ``man stand in dirt'' prone to appear at the same time. And we plot a bar chart in Figure~\ref{fig:web1} to show the probability of relationship hallucination when “man swings bat” occurs, the results show that frequently co-occurring relationships are more likely to result in hallucinations. And as shown in Figure~\ref{fig:co3}~(a), we show an image, with a person holding a hair dryer and does not use it to dry the hair, to the LLaVA, however, it answers that ``the person is drying her hair'' incorrectly. So, the related co-occurring relationships may also contribute to hallucination.

\textbf{Subject-Relationship Co-occurrence.}
There is usually a strong co-occurrence between the subject and certain behaviors in vision, for example, when the subject is a chef, he is usually cooking something instead of driving a car, as shown in Figure~\ref{fig:co}~(middle). This co-occurrence may lead to hallucinations, as shown in Figure~\ref{fig:web2} ``bus'' case. And as shown in Figure~\ref{fig:co3}~(b), the image is a cat wearing a soccer uniform with a soccer, when we ask LLaVA whether a ``soccer player'' kicking a ball, it answers ``Yes'' incorrectly. But when we ask it whether a ``cat'' kicking a ball, it can answer correctly. The relationship hallucinations resulting from this co-occurrence are widespread in LVLMs, due to the bias of the relevant training data towards the objects, as discussed in Relationship-Object Co-occurrence.

\textbf{Relationship-Object Co-occurrence.}
There is also a co-occurrence between the relationship and the object, and some high-frequency co-occurrence relationships, such as riding a bike, also often lead to hallucinations, as shown in Figure~\ref{fig:co} and Figure~\ref{fig:web3}. And as shown in Figure~\ref{fig:co3}~(c), it is hard for LVLM to distinguish between ``riding bike'' and ``walking bike'', but it is easy for it to distinguish whether the object the person is interacting with is a bike or a motorbike.
This may be related to the unbalanced distribution of data in the caption data, as shown in Table~\ref{tab:coco_sta} and Figure~\ref{fig:coco_cap}, the number of object words is much larger than that of relational words. Each caption contains an average of 3.6 object words, but only 1.3 relationship words. And 97,522 captions do not contain relationship words, while only 419 captions do not contain object words. Thus forming a long-tail distribution between the relationship words and the object words. 

\begin{table}[t]
\caption{The data static of relationship words and object words in COCO captions. The `Obj' and `Rel' denote the number of objects and relationship words. The `Spatial Rel' denotes the number of spatial relationship phrases whose proportion in relation words is about 1/5.}
\label{tab:coco_sta}
\vskip 0.15in
\begin{center}
\begin{small}
\begin{sc}
\begin{tabular}{cccc}
\toprule
Captions & Obj & Rel & Spatial Rel \\
\midrule
591,753    & 2,130,794 & 764,711 & 159,649 \\
\bottomrule
\end{tabular}
\end{sc}
\end{small}
\end{center}
\vskip -0.2in
\end{table}

\begin{figure*}[ht]
\vskip 0.2in
\begin{center}
\centerline{\includegraphics[width=0.96\textwidth]{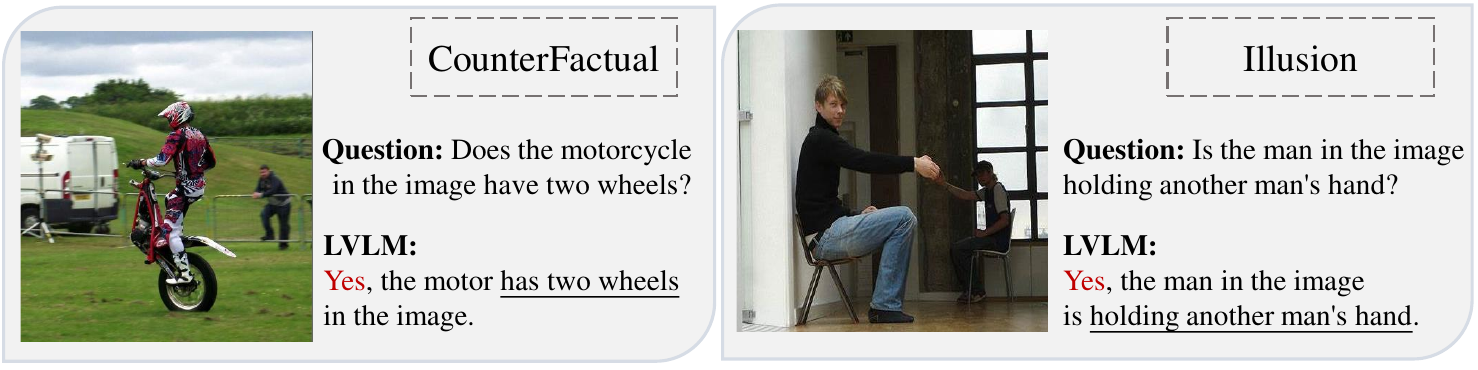}}
\caption{The examples of counterfactual relationship hallucination~(left) and illusion relationship hallucination~(right). The wrong answers are marked in red, and the relationships in the answers are underlined.}
\label{fig:web}
\end{center}
\vskip -0.3in
\end{figure*}

\begin{table}[t]
\caption{Results of LVLMs on the other relationship hallucinations. We here report counterfactual accuracy, illusion accuracy, and total accuracy due to all questions with label no.}
\label{tab:web}
\vskip 0.15in
\begin{center}
\begin{small}
\begin{sc}
\begin{tabular}{l|ccc}
\toprule
\multirow{3}*{Model} & \multicolumn{3}{c}{Accuracy} \\
\cline{2-4}
~ & \makecell*[c]{Counter\\Factual} & Illusion & Total \\
\midrule
 LLaVA-1.5    &  36.36 & 0.00 & 21.05 \\
InstrutBLIP &  9.09 & 25.00 & 15.79 \\
mPLUG-Owl2 & 27.27 & 0.00 & 15.79 \\
Qwen-Vl & 50.00 & 0.00 & 30.00 \\
\bottomrule
\end{tabular}
\end{sc}
\end{small}
\end{center}
\vskip -0.2in
\end{table}

\subsection{Other Relationship Hallucinations}
Most relationship hallucinations in R-Bench are linked to the three mentioned types of relationship co-occurrence, though some other relational hallucinations might be overlooked. Consequently, we further analyze additional relationship hallucinations in LVLMs, including counterfactual relationship hallucination and illusion relationship hallucination. We collect about 20 corresponding images from the web, and artificially form the questions with label no. We use accuracy as a metric to evaluate the LVLMs on these questions due to all the questions only with label no.

\textbf{Counterfactual Relationship Hallucination.}
Existing LVLMs consist of a pre-trained visual encoder and LLM, and the LLM usually contains enough common-sense knowledge. Therefore, we assess whether LVLMs ignore the actual visual content and answer the question based on this common-sense knowledge directly. The result is shown in Table~\ref{tab:web}. The Qwen-VL has the best performance, yet only has an accuracy of 50\%. As shown in Figure~\ref{fig:web} (left), we show an image with a motorcycle consisting of only one wheel to the LVLM, but it misjudges that the motorcycle in the image has two wheels, even though this is true in common sense. This result shows that existing LVLMs tend to ignore the actual visual content and answer questions directly based on the knowledge from the LLM.

\textbf{Illusion Relationship Hallucination.}
The ability to understand context-based spatial relationships demonstrates the reasoning ability of LVLMs, which can be evaluated through illusion images. The result is shown in Table~\ref{tab:web} and Figure~\ref{fig:web} (right). The best performance is 25\% accuracy of InstructBLIP. The example as shown in Figure~\ref{fig:web}, there is a spatial dislocation here in the two men, and one can clearly tell this based on spatial relationships and context, but the LVLM argues that there is an interaction between these two spatially dislocated people. This indicates that the existing LVLMs still cannot reason about spatial relationships based on context.

\section{Limitations}
Firstly, our work exclusively focuses on analyzing relationship hallucinations in LVLMs, with plans to implement migration strategies based on our findings in future research.
Secondly, our analysis covers only a subset of the causes behind relationship hallucinations. We anticipate future works to offer a more comprehensive and in-depth exploration of this phenomenon.
Thirdly, a lack of an effective classification standard hinders a fine-grained categorization of relationship hallucination questions. Improved categorization could enhance the analysis of LVLM hallucinations.
Fourthly, the manually filtered data may not be entirely free of noise due to omissions and biases in the filtering process.
Lastly, we did not evaluate certain powerful closed-source LVLMs, like GPT-4, due to budget constraints.

\section{Conclusion}
In this paper, we are committed to evaluating and analyzing relationship hallucinations in LVLMs. We propose a novel Relationship Hallucination Benchmark~(R-Bench) that includes image-level concentrate on relationship
existence and instance-level questions for evaluating local visual comprehension. We analyze the relationship hallucination in the existing LVLMs on R-Bench and reveal several important relationship hallucination phenomena present in existing LVLMs. We anticipate that our findings will serve as inspiration for the community to delve into solutions addressing the relationship hallucinations of LVLMs.

 
\section*{Acknowledgements}
This work was supported by National Key R\&D Program of China (No.2022ZD0118201), the National Science Fund for Distinguished Young Scholars (No.62025603), the National Natural Science Foundation of China (No. U21B2037, No. U22B2051, No. 62072389), the National Natural Science Fund for Young Scholars of China (No. 62302411), China Postdoctoral Science Foundation (No. 2023M732948), the Natural Science Foundation of Fujian Province of China (No.2021J01002,  No.2022J06001), and partially sponsored by CCF-NetEase ThunderFire Innovation Research Funding (NO. CCF-Netease 202301).

\section*{Impact Statement}
This paper presents work whose goal is to advance the field of Machine Learning. There are many potential societal consequences of our work, none which we feel must be specifically highlighted here.



\bibliography{example_paper}
\bibliographystyle{icml2024}

\newpage
\appendix
\onecolumn
\section{Implementation Detail of LVLMs}
We evaluated the latest versions of several popular LVLMs and chose the largest version model we could run as much as possible. During inference, we perform low-bit quantization on the LLM to avoid the out of GPU memory. We list the models we evaluated as well as the parameters of their core components in Table~\ref{tab:lvlm}.
\begin{table*}[h]
\caption{The specific parameters of evaluted LVLMs.}
\label{tab:lvlm}
\vskip 0.15in
\begin{center}
\begin{small}
\begin{sc}
\begin{tabular}{l|cc|cc}
\toprule
 Model & Vision Encoder & Parameters & LLM & Parameters \\
\midrule
 LLaVA-1.5    & ViT-L/14 & 0.4B & Vicuna-v1.5 & 13B \\
 InstrutBLIP & ViT-G/14  &	2.0B & Vicuna & 13B \\
 mPLUG-Owl2 &  ViT-L/14 & 0.4B & Llama-2-Chat & 7B \\
 Qwen-VL & ViT-G/14 & 2.0B & Qwen & 7B \\
\bottomrule
\end{tabular}
\end{sc}
\end{small}
\end{center}
\vskip -0.1in
\end{table*}

\section{Additional Results of LVLMs on our R-Bench}
Nocaps datasets include in-domain, near-domain and out-domain images. We provide additional results of LVLMs on our R-Bench under in-domain, near-domain and out-domain subsets. We randomly perform 1:1 positive and negative sampling for each subset 5 times, and average the results, as shown in Table~\ref{suptab:result}.
\begin{table*}[t]
\caption{Results of LVLMs on our R-Bench. 
We compute average scores of 5 random subsets, and each subset has 1:1 pos-neg questions. The `box' and `mask' denote types of instance-level questions with bounding box and mask respectively.}
\label{suptab:result}
\vskip 0.15in
\begin{center}
\begin{small}
\begin{sc}
\begin{tabular}{lllccc|c|c}
\toprule
Type & Subset & Model & Accuracy & Precision & Recall & F1 Score & Yes \\
\hline
\multirow{12}{*}{Image} & \multirow{4}{*}{In-Domain} & LLaVA-1.5 & 67.08 & 61.07 & 96.71 & 74.86 & 80.28 \\
& & InstructBLIP & 67.36 & 61.02 & 98.63 & 75.39 & 81.94 \\
& & mPLUG-OWL2 & 68.06 & 62.85 & 90.41 & 74.15 & 72.92 \\
& & Qwen-VL & 78.88 & 78.32 & 80.28 & 79.28 & 51.61 \\
\cline{2-8}
& \multirow{4}{*}{Near-Domain} & LLaVA-1.5 & 69.9 & 63.37 & 97.25 & 76.73 & 78.33 \\
& & InstructBLIP & 67.7 & 61.71 & 96.73 & 75.35 & 80.0 \\
& & mPLUG-OWL2 & 72.37 & 66.44 & 92.7 & 77.4 & 71.21 \\
& & Qwen-VL & 77.91 & 74.33 & 85.77 & 79.64 & 58.13 \\
\cline{2-8}
& \multirow{4}{*}{Out-Domain} & LLaVA-1.5 & 66.2 & 60.39 & 97.35 & 74.54 & 81.92 \\
& & InstructBLIP & 65.43 & 59.97 & 96.12 & 73.86 & 81.44 \\
& & mPLUG-OWL2 & 70.29 & 65.1 & 89.5 & 75.37 & 69.86 \\
& & Qwen-VL & 75.5 & 72.11 & 83.55 & 77.41 & 58.24 \\
\hline
\multirow{12}{*}{Box} & \multirow{4}{*}{In-Domain} & LLaVA-1.5 & 51.18 & 50.6 & 98.43 & 66.84 & 97.25 \\
& & InstructBLIP & 52.75 & 51.57 & 89.8 & 65.51 & 87.06 \\
& & mPLUG-OWL2 & 49.02 & 49.37 & 82.35 & 61.72 & 83.33 \\
& & Qwen-VL & 56.47 & 53.89 & 89.41 & 67.24 & 82.94 \\
\cline{2-8}
& \multirow{4}{*}{Near-Domain} & LLaVA-1.5 & 53.39 & 51.85 & 94.75 & 67.02 & 91.36 \\
& & InstructBLIP & 51.68 & 50.97 & 87.85 & 64.51 & 86.17 \\
& & mPLUG-OWL2 & 55.06 & 53.11 & 86.39 & 65.78 & 81.33 \\
& & Qwen-VL & 58.67 & 55.53 & 86.96 & 67.78 & 78.29 \\
\cline{2-8}
& \multirow{4}{*}{Out-Domain} & LLaVA-1.5 & 52.89 & 51.55 & 96.14 & 67.11 & 93.25 \\
& & InstructBLIP & 52.89 & 51.7 & 87.71 & 65.05 & 84.82 \\
& & mPLUG-OWL2 & 56.39 & 53.91 & 88.07 & 66.88 & 81.69 \\
& & Qwen-VL & 61.33 & 57.18 & 90.12 & 69.96 & 78.8 \\
\hline
\multirow{12}{*}{Mask} & \multirow{4}{*}{In-Domain} & LLaVA-1.5 & 55.29 & 52.89 & 96.86 & 68.42 & 91.57 \\
& & InstructBLIP & 55.29 & 53.15 & 89.02 & 66.55 & 83.73 \\
& & mPLUG-OWL2 & 53.73 & 52.34 & 81.96 & 63.87 & 78.24 \\
& & Qwen-VL & 61.76 & 59.05 & 76.47 & 66.62 & 64.71 \\
\cline{2-8}
& \multirow{4}{*}{Near-Domain} & LLaVA-1.5 & 54.46 & 52.46 & 95.00 & 67.60 & 90.54 \\
& & InstructBLIP & 56.01 & 53.87 & 83.54 & 65.50 & 77.53 \\
& & mPLUG-OWL2 & 57.53 & 55.04 & 82.15 & 65.92 & 74.62 \\
& & Qwen-VL & 58.39 & 56.38 & 74.05 & 64.01 & 65.66 \\
\cline{2-8}
& \multirow{4}{*}{Out-Domain} & LLaVA-1.5 & 54.64 & 52.54 & 96.02 & 67.92 & 91.39 \\
& & InstructBLIP & 56.63 & 54.29 & 83.73 & 65.87 & 77.11 \\
& & mPLUG-OWL2 & 55.24 & 53.51 & 79.76 & 64.05 & 74.52 \\
& & Qwen-VL & 63.49 & 60.33 & 78.8 & 68.33 & 65.3 \\
\bottomrule
\end{tabular}
\end{sc}
\end{small}
\end{center}
\vskip -0.1in
\end{table*}

\section{Impact of Color Selection.}
We select red and green as the base colors for marking because these colors are generally more distinguishable and easier for the model to recognize compared to other colors. Additionally, red and green tend to stand out more prominently within images, aiding in clearer object identification. To investigate the impact of color selection on model performance, we conducted experiments where we either swapped the red and green colors or replaced them with gold and pink, as shown in Ta. The results showed that interchanging red and green had only a minor effect on performance. However, substituting red and green with gold and pink led to various degrees of performance decline across different models, except for LLaVA, which demonstrated strong robustness to the change in colors.
\begin{table*}[t]
\caption{Impact of the color selection.}
\label{suptab:color}
\vskip 0.15in
\begin{center}
\begin{small}
\begin{sc}
\begin{tabular}{llccc|c|c}
\toprule
Type & Model & Accuracy & Precision & Recall & F1 Score & Yes \\
\toprule
 \multirow{4}{*}{Box}
         & LLaVA-1.5 & 53.15 & 51.71 & 95.53 & 67.10 & 92.37 \\
         & InstructBLIP & 51.95 & 51.14 & 87.39 & 64.52 & 85.44 \\
         & mPLUG-OWL2 & 53.90 & 52.38 & 85.67 & 65.01 & 81.77 \\
         & Qwen-VL & 58.82 & 55.56 & 88.17 & 68.16 & 79.35 \\
        \hline
        \multirow{4}{*}{Box (shift red and green)}
         & LLaVA-1.5 & 53.52 & 51.91 & 95.92 & 67.36 & 92.40 \\
         & InstructBLIP & 52.57 & 51.51 & 87.48 & 64.84 & 84.91 \\
         & mPLUG-OWL2 & 53.80 & 52.40 & 82.71 & 64.16 & 78.91 \\
         & Qwen-VL & 59.48 & 56.01 & 88.35 & 68.56 & 78.86 \\
        \hline
        \multirow{4}{*}{Box (gold and pink)}
         & LLaVA-1.5 & 53.38 & 51.81 & 96.79 & 67.49 & 93.41 \\
         & InstructBLIP & 52.22 & 51.35 & 84.38 & 63.85 & 82.16 \\
         & mPLUG-OWL2 & 52.97 & 51.92 & 80.02 & 62.98 & 77.05 \\
         & Qwen-VL & 57.97 & 55.14 & 85.55 & 67.06 & 77.58 \\
\bottomrule
\end{tabular}
\end{sc}
\end{small}
\end{center}
\vskip -0.1in
\end{table*}

\section{Examples of Instance-level Relationship Hallucination}
\begin{figure*}[ht]
\vskip 0.2in
\begin{center}
\centerline{\includegraphics[width=0.94\textwidth]{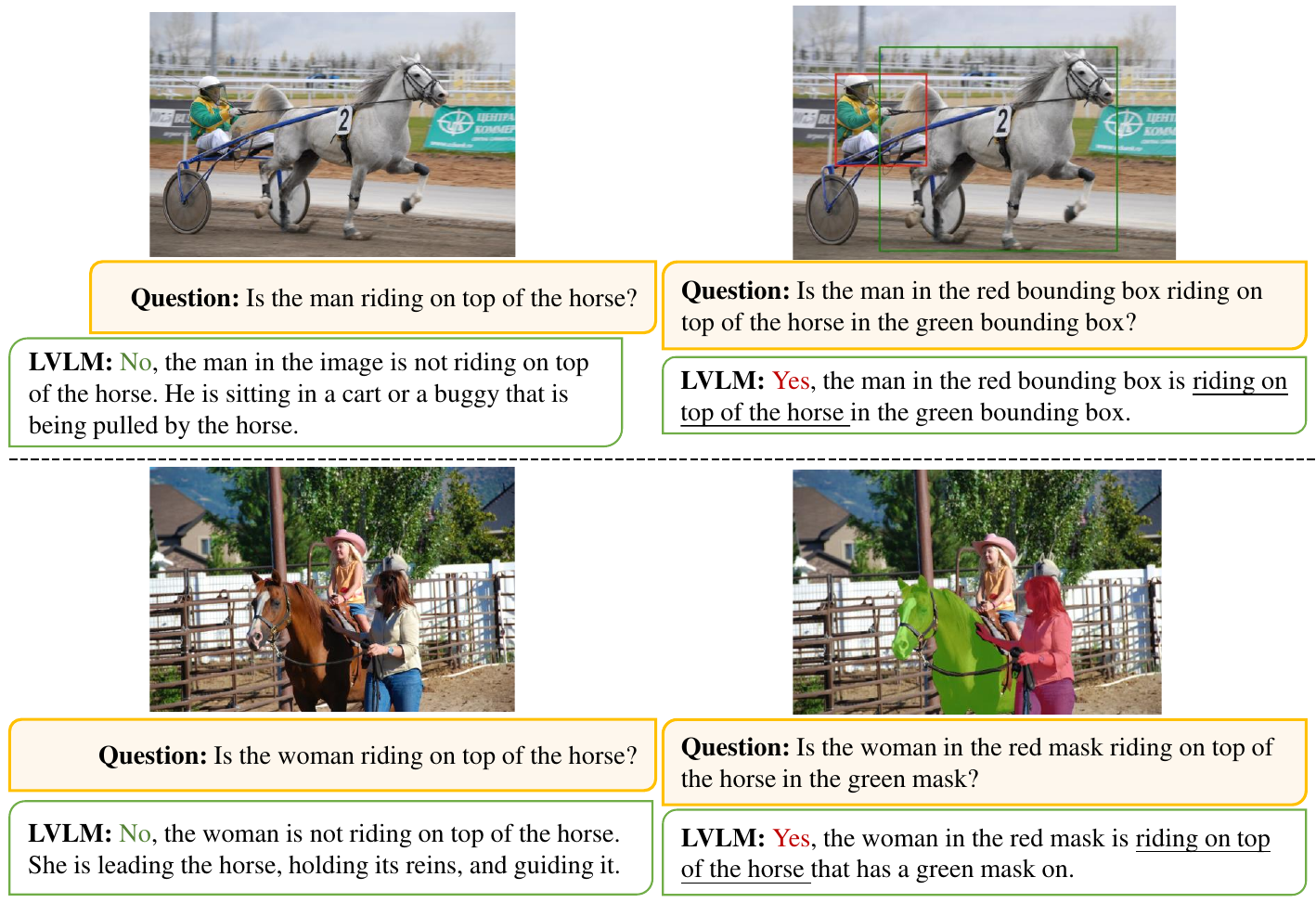}}
\caption{The examples of comparing between the image-level~(left) and instance-level~(right) relationship hallucination. The wrong answers are marked in red, and the correct answers are marked in green.}
\label{fig:ins}
\end{center}
\vskip -0.2in
\end{figure*}

We compare the image-level and instance-level relationship hallucination. As shown in Figure~\ref{fig:ins}, the LLaVA can better answer image-level questions than the instance-level ones. 

\section{Visualization Examples}
\begin{figure*}[ht!]
\vskip -0.1in
\begin{center}
\centerline{\includegraphics[width=0.93\textwidth]{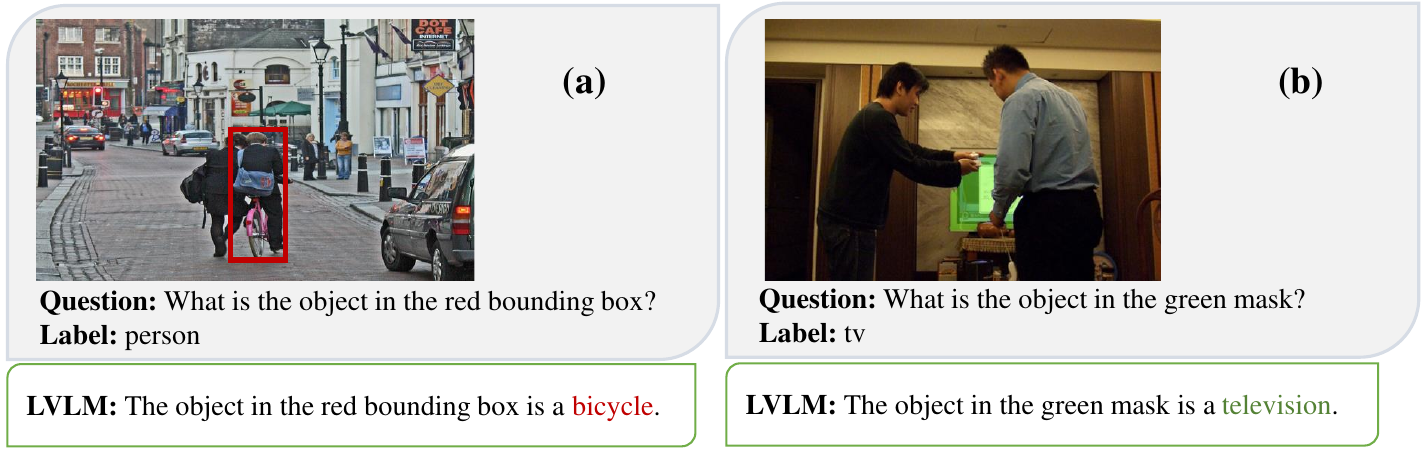}}
\caption{The examples of the reference object recognition base on bounding box~(left) and mask~(right). The wrong answers are marked in red, and the correct answers are marked in green. The mask provides more accurate reference than the box. And the LVLM generates the correct responses, with objects in different forms
but with correct meanings.}
\label{fig:ref_obj}
\end{center}
\vskip -0.2in
\end{figure*}

\begin{figure*}[ht!]
\vskip -0.1in
\begin{center}
\centerline{\includegraphics[width=0.97\textwidth]{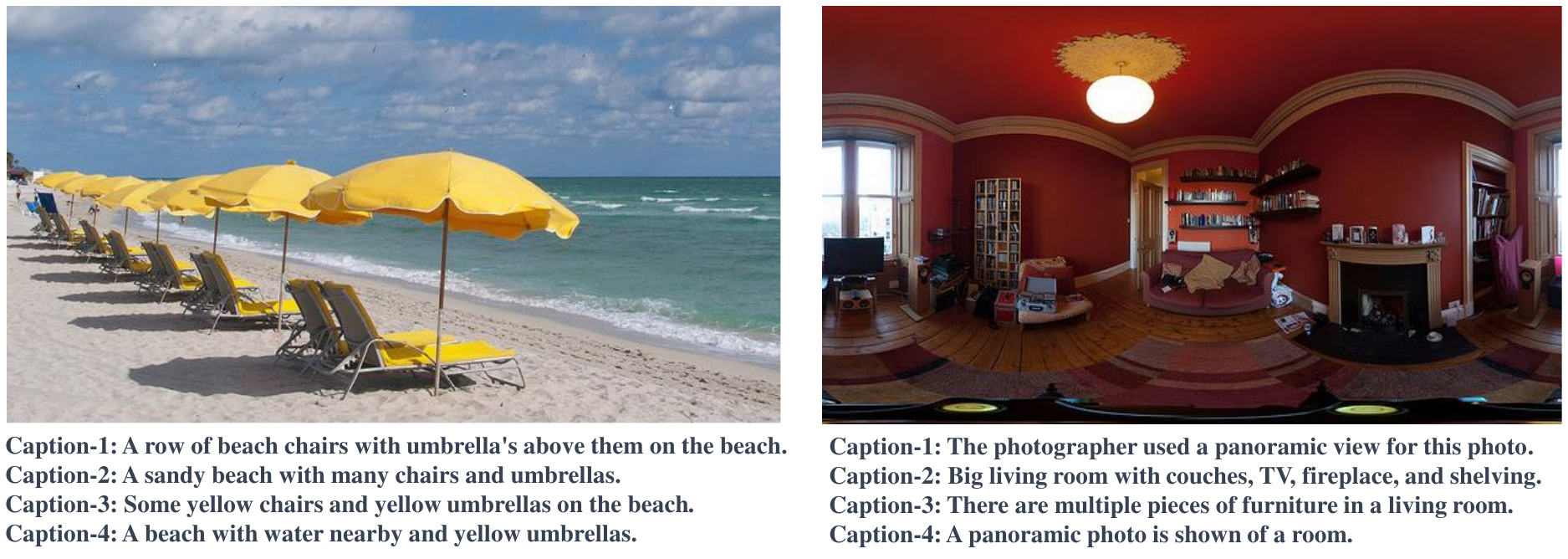}}
\caption{The examples of the coco image with captions. There are many image without any relationship descriptions or with unbalanced descriptions between the relationship and the objects.}
\label{fig:coco_cap}
\end{center}
\vskip -0.2in
\end{figure*}

\begin{figure*}[ht!]
\vskip -0.1in
\begin{center}
\centerline{\includegraphics[width=0.93\textwidth]{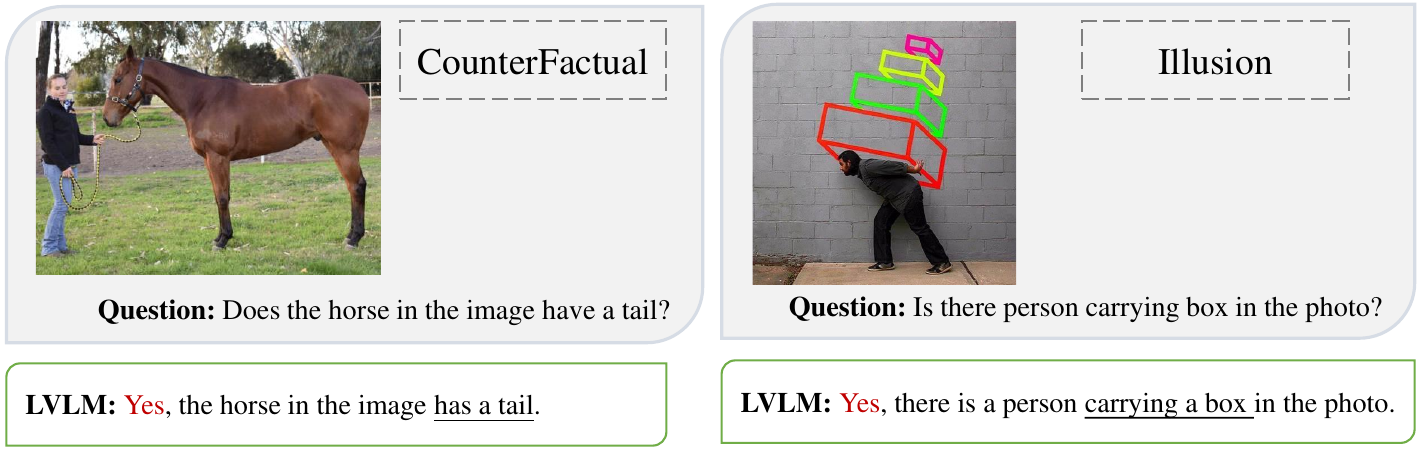}}
\caption{The other examples of counterfactual relationship hallucination and illusion relationship hallucination.}
\label{fig:web_sup}
\end{center}
\vskip -0.2in
\end{figure*}

There are some visualization examples, including the reference object recognition in Figure~\ref{fig:ref_obj}, the coco image with captions in Figure~\ref{fig:coco_cap} and the other examples of counterfactual relationship hallucination and illusion relationship hallucination in Figure~\ref{fig:web_sup}.

\section{Data Static}
\begin{figure*}[ht!]
\begin{center}
\centerline{\includegraphics[width=0.97\textwidth]{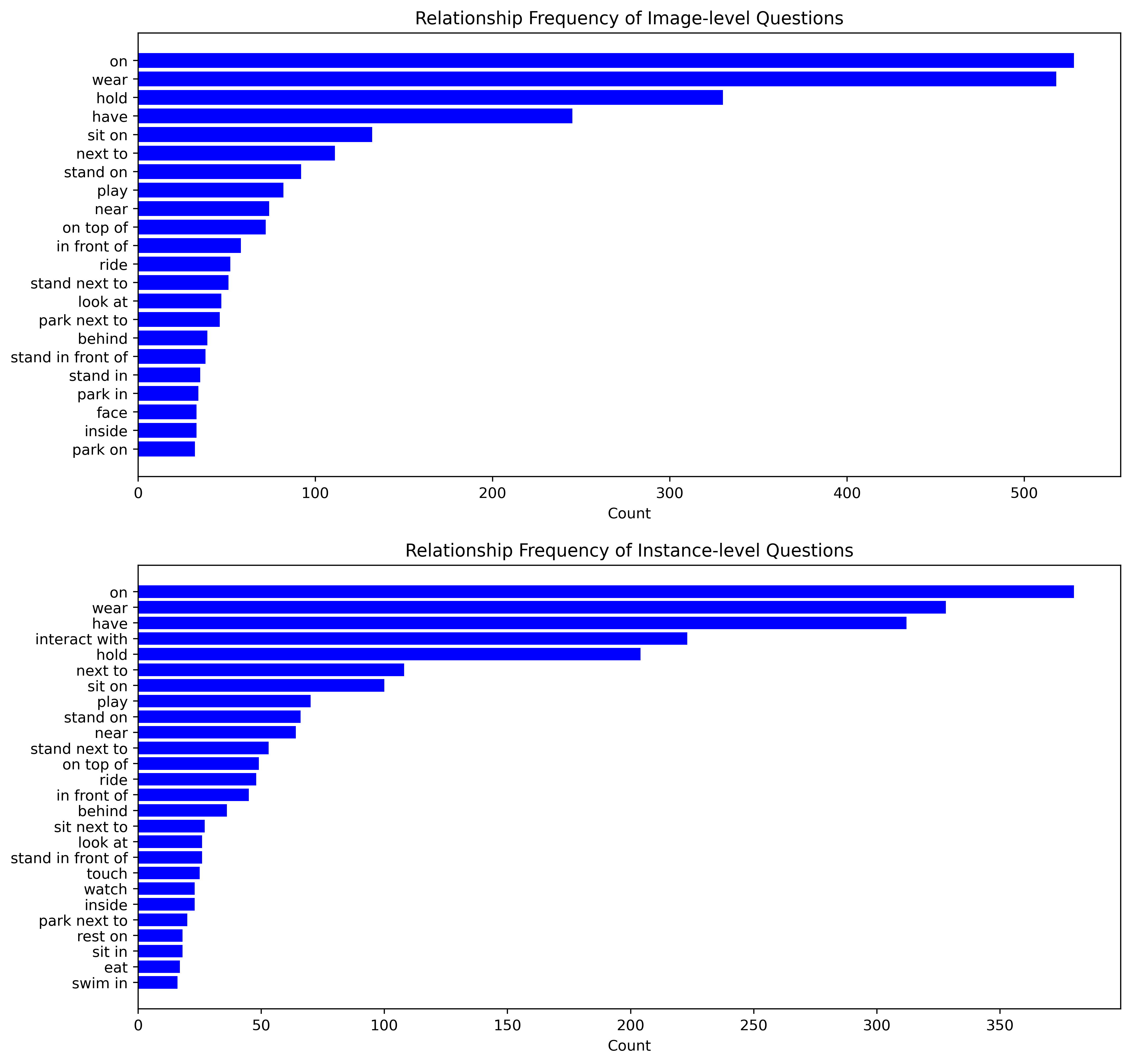}}
\vskip -1em
\caption{The relationship frequency in the image-level and instance-level questions.}
\label{fig:qs_bar}
\end{center}
\vskip -0.5em
\end{figure*}

\begin{figure*}[ht!]
\begin{center}
\centerline{\includegraphics[width=0.97\textwidth]{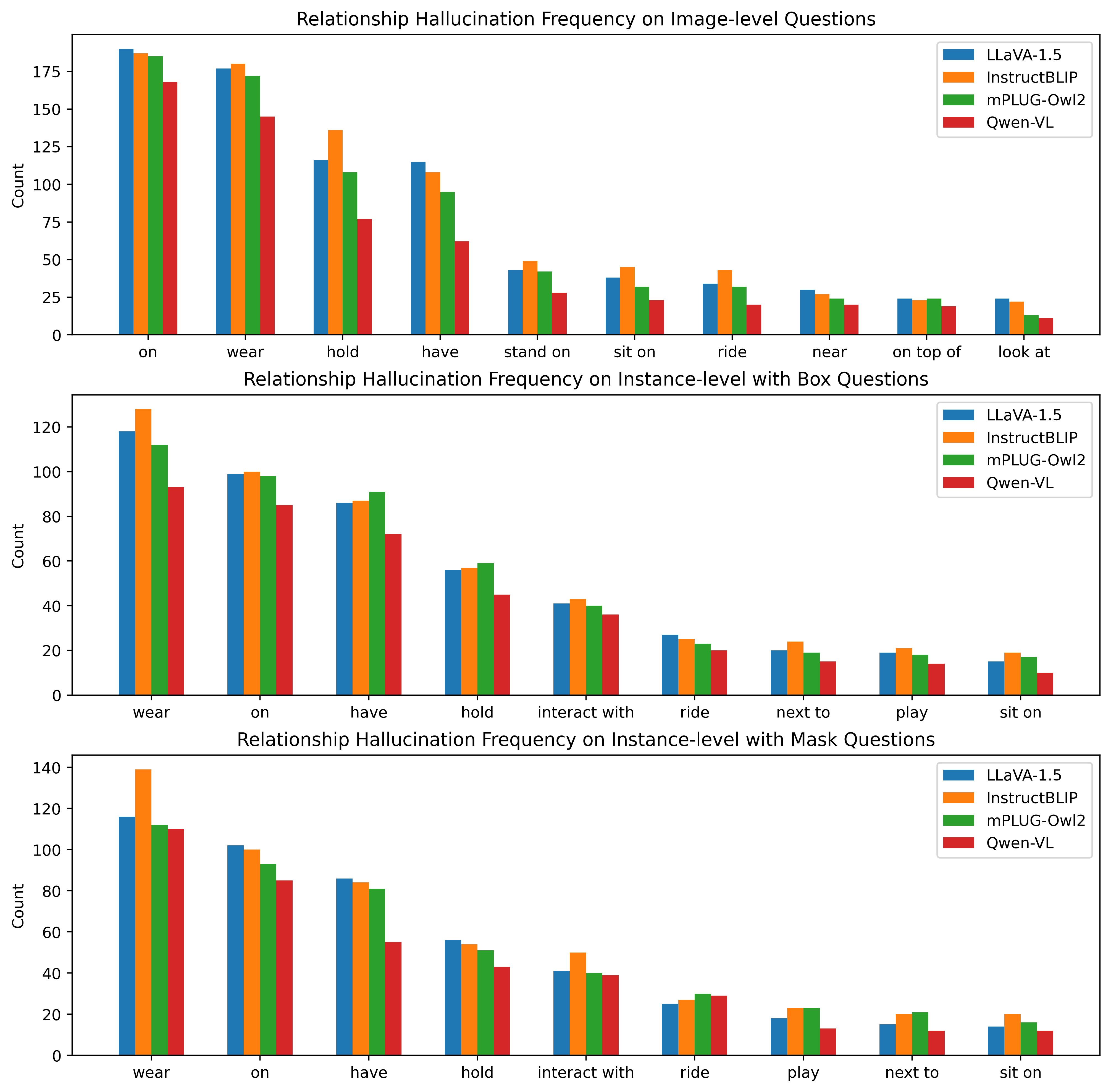}}
\vskip -1em
\caption{The relationship hallucination frequency of the LVLMs on the image-level and instance-level questions.}
\label{fig:hal_bar}
\end{center}
\vskip -0.5em
\end{figure*}

We show some of the high-frequency relationships that appear in R-Bench in Figure~\ref{fig:qs_bar}. And the high-frequency relationship hallucination of the LVLMs on the image-level and instance-level questions in Figure~\ref{fig:hal_bar}.

\end{document}